# A Transformer-based Math Language Model for Handwritten Math Expression Recognition

Huy Quang Ung [0000-0001-9238-8601], Cuong Tuan Nguyen [0000-0003-2556-9191], Hung Tuan Nguyen [0000-0003-4751-1302], Thanh-Nghia Truong [0000-0002-8635-8534], and Masaki Nakagawa [0000-0001-7872-156X]

Tokyo University of Agriculture and Technology, Tokyo, Japan

ungquanghuy93@gmail.com
fx4102@go.tuat.ac.jp
{ntuanhung, thanhnghiadk}@gmail.com
nakagawa@cc.tuat.ac.jp

**Abstract.** Handwritten mathematical expressions (HMEs) contain ambiguities in their interpretations, even for humans sometimes. Several math symbols are very similar in the writing style, such as dot and comma or "0", "O", and "o", which is a challenge for HME recognition systems to handle without using contextual information. To address this problem, this paper presents a Transformer-based Math Language Model (TMLM). Based on the self-attention mechanism, the high-level representation of an input token in a sequence of tokens is computed by how it is related to the previous tokens. Thus, TMLM can capture long dependencies and correlations among symbols and relations in a mathematical expression (ME). We trained the proposed language model using a corpus of approximately 70,000 LaTeX sequences provided in CROHME 2016. TMLM achieved the perplexity of 4.42, which outperformed the previous math language models, i.e., the *N*-gram and recurrent neural network-based language models. In addition, we combine TMLM into a stochastic context-free grammar-based HME recognition system using a weighting parameter to re-rank the top-10 best candidates. The expression rates on the testing sets of CROHME 2016 and CROHME 2019 were improved by 2.97 and 0.83 percentage points, respectively.

## 1    Introduction

Nowadays, devices such as pen-based or touch-based tablets and electronic whiteboards are becoming more and more popular for users as educational media. Learners can use them to learn courses and do exercises. Especially, educational units can use those devices to support their online learning platform in the context of the SARS-CoV-2 (COVID-19) widely spreading worldwide. These devices provide a user-friendly interface for learners to input handwritten mathematical expressions, which is more natural and quicker than common editors such as Microsoft Equation Editor, MathType, and LaTeX. Due to the demands of real applications, the research on handwritten mathematical expression (HME) recognition has been conceived as an important role in document analysis since the 1960s and very active during the last two decades. The performance of HME recognition systems has been significantly improved according to the series of competitions on recognition of handwritten mathematical expressions (CROHME) [1].

However, there remain challenging problems in HME recognition. One problem is that there are lots of ambiguities in the interpretation of HMEs. For instance, there exist math symbols that are very similar in the writing style, such as "0", "o", and "O" or dot and comma. These ambiguities challenge HME recognition without utilizing contextual information. In addition, recognition systems without using predefined grammar rules such as the encoder-decoder model [2, 3] might result in syntactically unacceptable misrecognitions. One promising solution for these problems is to combine an HME recognition system with a math language model. Employing language models for handwritten text recognition has shown effectiveness in previous research [4–6].

A mathematical expression (ME) has a 2D structure represented by several formats such as MathML, one-dimensional LaTeX sequences, and two-dimensional symbol layout trees [7]. Almost all recent HME recognition systems output their predictions as the LaTeX sequences since LaTeX is commonly used in real applications. Thus, we focus on an ME language model for the LaTeX sequences in this paper.

There are some common challenges in modeling MEs similar to natural language processing. First, there is a lack of corpora of MEs as MEs rarely appear in daily documents. Secondly, there are infinite combinations of symbols and spatial relationships in MEs. Thirdly, there are long-term dependencies and correlations among symbols and relations in an ME. For example, "(" and ")" are often used to contain a sub-expression, and if they contain a long sub-expression, it is challenging to learn the dependency between them.

There are several methods to modeling MEs. The statistical $N$-gram model was used in [8]. It assigns a probability for the $n$-th tokens given $(n-1)$ previous tokens based on the maximum likelihood estimation. However, the $N$-gram model might not represent the long dependencies due to the limitation of the context length. Increasing this length might lead to the problem of estimating a high-dimensional distribution, and it requires a sufficient amount of training corpus. In practical applications, the trigram model is usually used, and the 5-gram model is more effective when the training data is sufficient. The recurrent neural network-based language model (RNNLM) proposed by [9] was utilized in HME recognition systems [1, 2]. RNNLM predicts the $n$-th token given $(n-1)$ previous tokens in previous time steps. However, they still face the problem of the long-term dependencies.

In recent years, the transformer-based network using a self-attention mechanism has achieved impressive results in natural language processing (NLP). In the language modeling task, Al-Rfou et al. [10] presented a deep transformer-based language model (TLM) for character-level modeling and showed the effectiveness against RNNLM.

In this paper, we present the first transformer-based math language model (TMLM). Based on the self-attention mechanism, the high-level representation of an input token in a sequence of tokens is computed by how it is related to the previous tokens so that TMLM can capture long dependencies and correlations in MEs. Then, we propose a method to combine TMLM into a stochastic context-free grammar-based HME recognizer. In our experiments, we show that our TMLM outperforms the *N*-gram model and RNNLM in the task of modeling MEs.

The rest of the paper is organized as follows. Section 2 briefly presents related research. Section 3 describes our method in detail. Section 4 presents our experiments for evaluating the proposed method. Finally, section 5 concludes our work and discusses future works.

## 2    Related works

Language models are well-known as generative models and autoregressive models since they predict the next state of a variable given its previous states. In NLP, Radford et al. [11, 12] proposed Generative Pre-Training models (GPT and GPT-2) with high achievements on NLP benchmarks based on the vanilla transformer-based network in [13]. Their models are trained by the casual language modeling loss, then fine-tuned for multitask learning such as text classification, question answering, and similarity. Dai et al. [14] presented a Transformer-XL for capturing extended length of context using a recurrent architecture for context segments. Transformer-XL can learn dependency that is 80% longer than RNNs, 450% longer than TLM. The inference speed is 1,800 times faster than TLM by caching and reusing previous computations. XLNet presented by Yang et al. [15], is the first model utilizing bidirectional contexts for transformer-based language models. This model significantly outperformed the conventional BERT model [16] in 20 tasks of NLP benchmarks.

There are several studies combining HME recognition systems with pre-trained language models. Wu et al. [8] combined their encoder-decoder HME recognizer with a pre-trained 4-gram model to get the *N* best paths. Zhang et al. [2] utilized a Gated Recurrent Unit-based language model (GRULM) for their HME recognizer that is an encoder-decoder model with temporal attention. This attention is to help the decoder determine the reliability of spatial attention and that of the language model per time step. The language models improved the expression rate by around 1 percentage point. Hence, the approach for combining language models into recognition systems is essential to study.

In CROHME 2019 [1], the Samsung R&D team used a probabilistic context-free grammar-based recognizer combined with two bigram language models, i.e., a language sequence model and a language model for spatial relationships. Besides, the MyScript team used LSTM-based language models for their grammar-based recognition system.

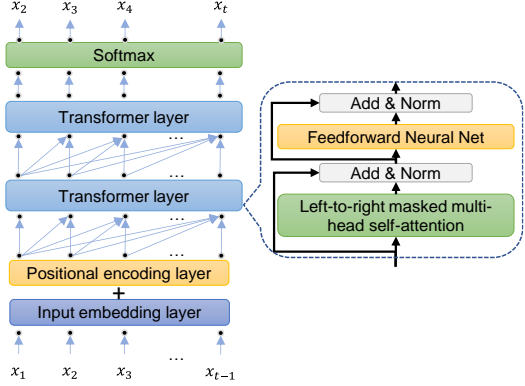

**Fig. 1.** Overview of the proposed transformer-based language model
with two transformer layers.

## 3    Proposed method

Given a sequence of tokens $X = (x_1, x_2 \dots, x_N)$, constructing a language model is to estimate the joint probability $P(X)$, which is often auto-regressively factorized as $P(X) = \prod_t P(x_t | X_{<t})$ where $X_{<t} = (x_1, \dots, x_{t-1})$. According to this factorization, the problem reduces to estimating each conditional factor $P(x_t | X_{<t})$. In this paper, our proposed model with a self-attention mechanism encodes the context $X_{<t}$ to produce the categorical probability of the token $x_t$.

In this section, we first describe our proposed TMLM, which is mainly based on [10]. Then, we present a method for combining our model with an HME recognizer.

### 3.1    Transformer-based math language model

TMLM consists of three main parts: an input embedding layer, a positional encoding layer (PE), and a stack of transformer layers, as shown in Fig. 1. First, sequential input tokens $\{x_1, x_2, \dots, x_N\}$ are fed into the input embedding to embed the categories of discrete tokens into a continuous space for better representation. Secondly, each embedded vector according to each input token is added by a PE vector to present the token's position in the sequence. The detail of the PE is presented later in this section. Thirdly, the outputs of the input embedding and positional encoding are passed into stacked transformer layers to learn high-level representation based on the self-attention mechanism. Finally, the output of the top transformer layer is input to a softmax layer to obtain the categorical probability for the token $x_t$ given $\{x_1, \dots, x_{t-1}\}$. Although all input tokens are fed into our model at the same time, the model is restricted to attend only tokens on the left side of $x_t$ to produce $P(x_t|x_1, \ldots, x_{t-1})$ by a mask in the transformer layer.

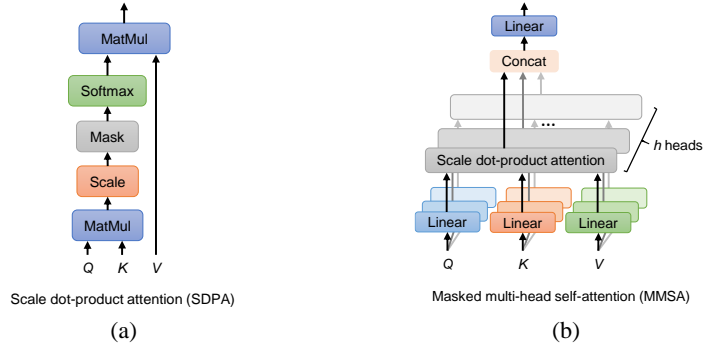

**Fig. 2.** Illustration of scale dot-product attention and masked multi-head attention.

The architecture of the transformer layer is based on the decoder of the conventional transformer-based model [16]. It consists of a masked multi-head self-attention (MMSA), layer normalization [17], and a feedforward neural network, as shown in Fig. 1. In addition, residual connections are added for the model to learn better. Here, we present MMSA and PE, which play important roles in our model.

**Masked multi-head self-attention.** This layer receives the representation of input tokens and outputs the higher representation for the tokens based on how each token is related to others. MMSA includes multiple attention functions, which allow the model to attend information from different representation subspace. We firstly present a masked single-head self-attention.

A traditional attention function can be described as the mapping of a query and a set of key-value pairs to produce an output. Note that the query, the keys, and the values are all vectors. The output is a weighted sum of the values, where the weight assigned to each value is computed by a compatibility function of the query with the corresponding key of the value.

The masked single-head self-attention function, called scaled dot-product attention (SDPA), are also based on the queries ($Q$), the keys ($K$) of dimension $d_k$, and the values ($V$) of dimension $d_v$ as shown in Fig. 2(a). We compute the dot products of the query with all keys, then scale them by $\sqrt{d_k}$. Next, we apply a mask to restrict the model to attend only the left side of the current predicted token. We then apply a softmax function to obtain the weights on the values. The output of this attention function is formulated as follows:

$$Att(Q, K, V) = \text{softmax}\left(\frac{QK^T}{\sqrt{d_k}}\right)V \qquad (1)$$

SDPA is called a "head" in MMSA. The architecture of MMSA including $h$ heads is shown in Fig. 2(b). With multiple heads, we project the queries, keys, and values $h$ times with three different learnable linear projections. On each of these projected versions of queries, keys, and values, we then perform SDPAs in parallel. Then, we concatenate their outputs and once again project to obtain the final output of MMSA.

**Positional encoding.** Since tokens $(x_1, x_2, \ldots, x_N)$ are input to our model at the same time and there is no convolutional/recurrent layer, the model cannot exploit the positional information of tokens. It is a serious problem for the task of language modeling. To address it, we utilize PE having the same dimensionality as the input embedded vector, $R^{N \times d_{embed}}$ ($d_{embed}$ is the dimension of the input embedded vector). Then, we add PE to the input embedded vector to provide the positional information for our model. PE of the $p$-th token and the $i$-th dimension is computed by the sine and cosine function as follows:

$$PE(p, i) = \begin{cases} \sin\left(\dfrac{p}{10000^{i/d_{embed}}}\right) & \text{if } i \text{ is even} \\ \cos\left(\dfrac{p}{10000^{(i-1)/d_{embed}}}\right) & \text{otherwise} \end{cases} \quad (2)$$

### 3.2 Combining language model with HME recognizer

In this study, we use a language model to sort the top-$M$ best candidates outputted from the stochastic context-free grammar-based HME recognizer. Given $M$ candidates $\{c_1, c_2, \ldots, c_M\}$ of LaTeX sequences and their corresponding scores, the combined scores are computed as follows:

$$Score_{comb}(c_i) = Score_{recog}(c_i) + \alpha \times Score_{LM}(c_i) \quad (3)$$

where $Score_{recog}(c_i)$ and $Score_{LM}(c_i)$ are the scores of the $i$-th candidate, $c_i$, from the HME recognizer and the language model, respectively. $Score_{comb}(c_i)$ is the combined score of $c_i$. $\alpha$ is a weighting parameter to balance between recognition and language scores. Note that $Score_{LM}(c_i)$ is the sum of logarithms of conditional probabilities output from the language model and normalized by the length of the candidate, $c_i$. For this combination method, we refer to the HME recognizer producing $Score_{recog}(c_i)$ based on the sum of logarithms of probability terms. The candidate having the highest combined score is the final recognition result.

## 4 Experiments

This section presents evaluations of our proposed TMLM using a corpus of LaTeX sequences. We also evaluate TMLM when combining with the HME recognizer proposed in [18] and present error analyses.

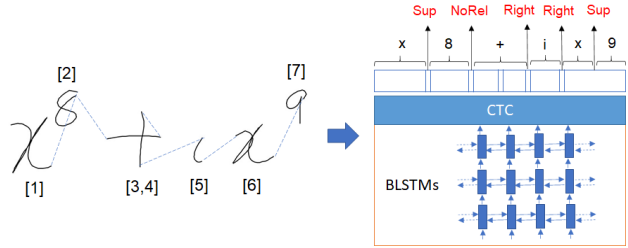

**Fig. 3.** Illustration for symbol-relation temporal classifier.

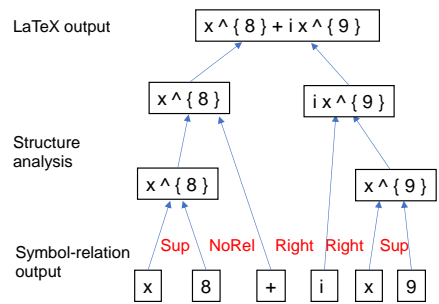

**Fig. 4.** Symbol-level parser.

### 4.1 Dataset

We uses a corpus of 68,862 LaTeX sequences provided in CROHME 2016 [19]. For preprocessing steps, we first filtered invalid syntax LaTeX sequences and removed style-related characters such as "\mathrm", "\textrm", and so on. Then, we normalized the LaTeX sequences into the same format as the output of our HME recognizer. For example, "{a}^{2}" is normalized as "a^{2}". The corpus is partitioned into a training set, a validation set, and a testing set according to the ratio of 8:1:1. The number of symbols in the dictionary is 108, including the padding "<pad>" and the end-of-sequence symbol "<eos>".

### 4.2 HME recognizer

In this section, we present the online HME recognizer [18] used in our experiments. The recognizer receives a sequence of point-based features extracted from an input HME and outputs recognition result as a LaTeX sequence. It consists of two main stages: (1) A symbol-relation temporal classifier (SRTC) for segmenting and classifying symbols and spatial relationships in an HME and (2) A symbol-level parser (SLP). We denoted this HME recognizer as SRTC_SLP.

**Symbol-relation temporal classifier.** SRTC consists of three stacked Bidirectional Long-Short Term Memory (BLSTM) layers and a Connectionist Temporal Classification (CTC) layer at the top, as shown in Fig. 3. Its input is a sequence of point-based features, including the representation of off-strokes (pen movements between strokes). Here, we ideally assume that there are no delayed strokes in the input HMEs. The stacked multiple BLSTM layers encode bidirectional context from the input and learn high-level representation. Then, the CTC layer generates a sequence of symbols and spatial relationships. There are 7 types of spatial relationships: superscript, subscript, upper, lower, horizontal, inside, and no relation (denoted as "NoRel").

**Symbol-level parser.** Given the output of SRTC, SLP based on the Cocke–Younger–Kasami (CYK) algorithm [20] is applied to merge recognized symbols and spatial relationships along with predefined grammar rules, as shown in Fig. 4. This bottom-up method considers many possible combinations of hypotheses at the intermediate levels. Hence, it produces several candidates at the top of the combination tree even if the less promising candidates are pruned. Each candidate has a corresponding score computed based on the classification probabilities of symbols and spatial relationships.

**Training and testing.** SRTC_SLP was trained on the CROHME 2016 and CROHME 2019 training sets and tested on the CROHME 2016 and CROHME 2019 testing sets, respectively. Without using a language model, it achieved the expression rate of 53.44% and 52.38% on the CROHME 2016 and CROHME 2019 testing sets, respectively. This expression rate is higher than that of the state-of-the-art TAP recognizer (without using language models and/or ensemble methods) [2] by 3.22 percentage points in the expression rate on the CROHME 2016 testing set.

### 4.3 Experimental settings

In this section, we present settings for training the proposed TMLM. We evaluated TMLM with different numbers of transformer layers. The number of heads is fixed to 4 heads. The dimension of each head is set to 16. The context length of TMLM is fixed to 256, which covers the maximum sequence length in our LaTeX corpus. The dimension of vectors of input embedding and the dimension of hidden states are set to 256 and 512, respectively. The number of hidden nodes in the feedforward neural net layer is set to 1024. The dropout rate is set to 0.1. We applied an adaptive log-softmax function proposed in [21]. Our model is trained by the AdamW optimizer [22] with a learning rate of $10^{-5}$. The model is implemented based on the "Hugging Faces" library [23]. For combining language models with SRTC_SLP, we determined the parameter $\alpha \in R^+$ in Eq. 3 by applying the enumeration method on $\{0, 0.1, 0.2, ..., 2.0\}$. The chosen $\alpha$ parameters achieved the best expression rates on the CROHME 2014 testing set.

To evaluate language models, we utilized the perplexity measurement. Given a sequence of tokens $X = (x_1, x_2, ..., x_N)$, the perplexity of $X$ is the exponentiated average negative log-likelihood formulated as follows:

$$ppl(X) = \exp\left\{-\frac{1}{N}\sum_{t=1}^{N}\log p(x_t|X_{<t})\right\} \tag{4}$$

where $p(x_t|X_{<t})$ is the conditional probability outputted from the language model.

### 4.4 Evaluation

In this section, we compare the proposed language model with the previous methods. Then, we conduct experiments to compare the performance of those models when combined with the SRTC_SLP recognizer.

**Comparisons with other language modeling methods.** We compared our TMLM with the traditional $N$-gram model and GRULM. We increased the context length $N$ in the $N$-gram model to 11 since TMLM can attend all the past contexts for a fair comparison. For GRULM, we increased the number of GRU layers up to 3 layers for evaluating the performance as well as comparing with TMLM in the condition of a similar number of trainable parameters. The dimension of an input embedded vector and hidden states in GRULM are set as the same as in TMLM.

**Table 1.** Comparisons with other language modeling methods.

| Model | #Layers in model | #parameters | Perplexity |
|---|---|---|---|
| 3-grams | - | - | 9.603 |
| 5-grams | - | - | 7.557 |
| 9-grams | - | - | 6.550 |
| 11-grams | - | - | 6.500 |
| GRULM_1L | 1 | 1.3M | 6.050 |
| GRULM _2L | 2 | 2.8M | 6.049 |
| GRULM _3L | 3 | 4.4M | 6.377 |
| **Ours: TMLM_2L** | 2 | 2.7M | 4.598 |
| **Ours: TMLM_5L** | 5 | 6.3M | 4.509 |
| **Ours: TMLM_8L** | **8** | **10M** | **4.420** |

Table 1 presents the perplexity of the models on the testing set extracted from our LaTeX corpus, as mentioned in section 4.3. The results show that our proposed TMLM models outperform all $N$-gram models and all GRULMs, even using fewer trainable parameters. For the $N$-gram models, increasing the context length can improve the perplexity, but it seems to converge when $N$ reaches 11. GRULMs perform better than the $N$-gram models. Among GRULMs, the perplexity of GRULM_2L achieves the best, which implies that increasing the number of GRU layers is not adequate. On the other hand, TMLMs can learn better when increasing the number of transformer layers.

With nearly the same number of trainable parameters, TMLM_2L performs significantly better than GRULM_2L. It implies that the architecture of TMLM is much more effective than the traditional GRULM on modeling MEs.

**Evaluation on combining language models into the HME recognizer.** We combined the SRTC_SLP recognizer with the language models that achieved the best performance in the previous experiment (i.e., 11-grams, GRULM_2L, and TMLM_8L). In detail, the combined score in Eq. 3 is computed for the top-10 best candidates from SRTC_SLP.

**Table 2.** Expression rates on combining the HME recognizers with language models.

| Recognition system | Expression rate (%) | |
|---|---|---|
| | CROHME 2016 | CROHME 2019 |
| SRTC_SLP | 53.44 | 52.38 |
| SRTC_SLP + 11-grams | 56.15 | 52.54 |
| SRTC_SLP + GRULM_2L | 55.36 | 52.88 |
| **(Ours) SRTC_SLP + TMLM_8L** | **56.41** | **53.21** |
| (Zhang et al. [2]) TAP | 49.29 | - |
| (Zhang et al. [2]) TAP + GRUs | 50.41 | - |
| (Wu et al. [8]) PAL-v2 | 49.00 | - |
| (Wu et al. [8]) PAL-v2 + 4-grams | 49.35 | - |

LM: language model

**Table 3.** Percentages of corrected, miscorrected, and unchanged recognition results when combining the SRTC_SLP recognizer with language models.

| Dataset | Method | Corrected (%) | Miscorrected (%) | Unchanged (%) |
|---|---|---|---|---|
| CROHME 2016 | 11-grams | 4.01 | 1.31 | 94.68 |
| | GRULM_2L | 2.96 | **1.05** | 95.99 |
| | **TMLM_8L** | **4.62** | 1.66 | 93.72 |
| CROHME 2019 | 11-grams | 1.83 | 1.67 | 96.50 |
| | GRULM_2L | 1.92 | **1.42** | 96.66 |
| | **TMLM_8L** | **2.50** | 1.67 | 95.83 |

Table 2 presents the expression rates of the combined recognizers on the CROHME 2016 and CROHME 2019 testing sets. The combination of SRTC_SLP and TMLM_8L achieves the best expression rates in both testing sets. TMLM_8L improves 2.97 and 0.83 percentage points of the expression rates on the CROHME 2016 and CROHME 2019 testing set, respectively. The SRTC_SLP + 11-grams is better than SRTC_SLP + GRULM_2L in the CROHME 2016 testing set, but that result is opposed in the CROHME 2019 testing set.

Table 2 also presents the expression rates of the state-of-the-art HME recognizers that utilized math language models, i.e., TAP [2] and PAL_v2 [8]. Compared to those models, SRTC_SLP combined with our TMLM_8L yields the best expression rate on the CROHME 2016 testing set. Combining the language models only improved around 1 percentage point in the case of TAP and PAL_v2 while TMLM_8L improves 2.97 percentage points. Here, we cannot conclude that our method for utilizing a math language model is better than their methods since they utilized different types of HME recognizers as well as different LaTeX corpora to train their language models. We consider conducting more experiments on the combination method as a remaining work.

Table 3 shows the recognition results in more detail about the percentage of corrected cases, miscorrected cases, and unchanged cases when combining SRTC_SLP with three different language models on the CROHME 2016 and CROHME 2019 testing sets. The percentages of the corrected cases by TMLM_8L are the highest compared to 11-grams and GRULM_2L on both testing sets. However, that of the miscorrected cases by TMLM_8L is the worst compared to others. GRULM_2L caused the least miscorrections compared to others, but it could not correct many cases. 11-grams and TMLM_8L have comparable percentages of miscorrected cases, but TMLM_8L corrected more cases than 11-grams did.

### 4.5 Error analysis

In this section, we present some samples which are corrected or miscorrected when applying TMLM_8L with the SRTC_SLP recognizer.

Fig. 5(a) and Fig. 5(b) show two corrected cases. In Fig. 5(a), "$\alpha$" in the HME sample are recognized as "2" without using TMLM_8L since it seems to be similarly written as "2". However, the language score of the candidate with "$\alpha$" is significantly higher than the one with "2". Therefore, the recognizer combined TMLM_8L results in the correct prediction. TMLM_8L performs well since "$\alpha$" seems more likely to appear next to the trigonometry function (e.g., sine, cosine, and tangent) than a number. Similarly, "9" in the HME sample of Fig. 5(b) are correctly recognized by TMLM_8L.

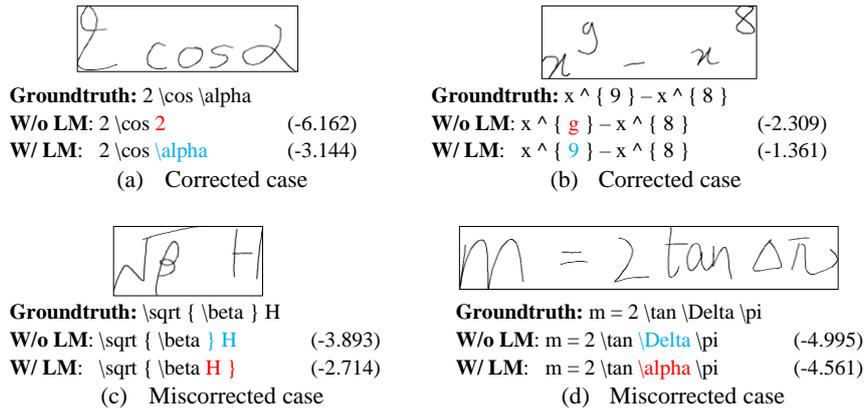

**Groundtruth:** 2 \cos \alpha
**W/o LM:** 2 \cos 2      (-6.162)
**W/ LM:** 2 \cos \alpha      (-3.144)

(a)    Corrected case

**Groundtruth:** x ^ { 9 } – x ^ { 8 }
**W/o LM:** x ^ { g } – x ^ { 8 }      (-2.309)
**W/ LM:** x ^ { 9 } – x ^ { 8 }      (-1.361)

(b)    Corrected case

**Groundtruth:** \sqrt { \beta } H
**W/o LM:** \sqrt { \beta } H      (-3.893)
**W/ LM:** \sqrt { \beta H }      (-2.714)

(c)    Miscorrected case

**Groundtruth:** m = 2 \tan \Delta \pi
**W/o LM:** m = 2 \tan \Delta \pi      (-4.995)
**W/ LM:** m = 2 \tan \alpha \pi      (-4.561)

(d)    Miscorrected case

**Fig. 5.** Examples of corrected and miscorrected cases when combining the SRTC_SLP recognizer and TMLM_8L (LM: language model). Each case shows an HME image, its ground truth, and its recognition candidates with/without TMLM_8L and their corresponding scores from TMLM_8L.

Fig. 5(c) and Fig. 5(d) show two miscorrected cases. The case in Fig. 5(c) is miscorrected since the language model score of the incorrect result is higher than that of the correct result. We can realize that the context, in this case, is not clear. The case in Fig. 5(d) is miscorrected since "$\alpha$" seems more likely to appear next to the tangent symbol than "$\Delta$".

According to those examples, we can see that modeling MEs is still challenging since the context in an ME is not clear and our corpus of MEs might not be enough to estimate the distribution of MEs.

## 5    Conclusion and future works

This paper presented a transformer-based math language model (TMLM) for improving the recognition rate of HME recognition systems. We showed that our TMLMs perform better than the traditional language models for MEs, i.e., the $N$-gram and GRULM. The best perplexity achieved is 4.42, resulted from TMLM_8L of 8 transformer layers. Combining TMLM_8L with the online HME recognizer in [18] improved the expression rate by 2.97 and 0.83 percentage points on the CROHME 2016 and CROHME 2019 testing set, respectively.

There are several remaining works. Firstly, we should enrich the source of ME LaTeX by collecting open sources on the internet. Secondly, we should modify our TMLM to exploit the bidirectional context in MEs. Thirdly, the method for jointly training an HME recognizer and a math language model should be studied for better optimization.

## Acknowledgement

This research is being partially supported by the grant-in-aid for scientific research (A) 19H01117 and that for Early Career Research 21K17761.